\pdfoutput=1

\documentclass[11pt]{article}

\usepackage{emnlp2022}

\usepackage{times}
\usepackage{tabularx}
\usepackage{longtable}
\usepackage{array}
\usepackage{booktabs}
\usepackage{xtab}
\usepackage{threeparttablex}
\usepackage{latexsym}
\usepackage{nicematrix,tikz}

\usepackage[T1]{fontenc}

\usepackage[utf8]{inputenc}
\usepackage{microtype}

\usepackage[color=green]{todonotes}

\usepackage{verbatim}
\usepackage{graphicx}

\mathchardef\mhyphen="2D
\usepackage{xcolor}
\usepackage{enumitem}
\usepackage{tcolorbox}

\newtcolorbox{mybox}[2][]
{
	colframe = #2!25,
	colback  = #2!10,
	coltitle = #2!20!black, 
	#1,
}

\usepackage{svg}
\usepackage{float}
\usepackage{hyperref}
\usepackage{amsmath}
\usepackage{amssymb} 
\usepackage{booktabs}
\usepackage{float}
\usepackage{xspace}
\usepackage{booktabs}
\usepackage{multirow}
\usepackage{multicol}

\usepackage{siunitx}

\newcommand\cont{\textsc{control}}
\newcommand\temp{\textsc{temporal}}
\newcommand\prog{\textsc{progressive}}

\newcommand\sandy{\textit{Sandy}}
\newcommand\clex{\textit{T26}}
\newcommand\humaid{\textit{Humaid}}

\definecolor{tablegray}{rgb}{0.2,0.2,0.2}

\title{The challenges of temporal alignment on Twitter during crises}

 \author{Aniket Pramanick,
 Tilman Beck,
 Kevin Stowe \and
 Iryna Gurevych \\
 Ubiquitous Knowledge Processing Lab (UKP Lab) \\
 Department of Computer Science and Hessian Center for AI (hessian.AI) \\
 Technical University Darmstadt \\
 \texttt{\url{www.ukp.tu-darmstadt.de}}
}

\begin{document}
\maketitle

\begin{abstract}

	Language use changes over time, and this impacts the effectiveness of NLP systems. 
	This phenomenon is even more prevalent in social media data during crisis events where meaning and frequency of word usage may change over the course of days. 
	Contextual language models fail to adapt temporally, emphasizing the need for temporal adaptation in models which need to be deployed over an extended period of time. 
	While existing approaches consider data spanning large periods of time (from years to decades), shorter time spans are critical for crisis data.
	We quantify temporal degradation for this scenario and propose methods to cope with performance loss by leveraging techniques from domain adaptation.
	To the best of our knowledge, this is the first effort to explore effects of rapid language change driven by adversarial adaptations, particularly during natural and human-induced disasters.
	Through extensive experimentation on diverse crisis datasets, we analyze under what conditions our approaches outperform strong baselines while highlighting the current limitations of temporal adaptation methods in scenarios where access to unlabeled data is scarce. \footnote{We publish the code for our experiments at \url{https://github.com/UKPLab/emnlp2022-temporal-adaptation}.}
	
\end{abstract}

\section{Introduction}

Patterns of language use change constantly over time, often in predictable and analyzable ways \cite{hamilton2016, kulkarni2015, sommerauer2019}. 
As language changes, the performance of NLP systems can be negatively impacted \cite{lazaridou2021mind}. 
In most scenarios, training corpora are derived from a snapshot of data at some moment of time in the past, which puts the reliability of model performance on future data into question.
Yet, there lacks a concrete reasoning or evidence that temporal adaptation elevates model performance. 
Despite the popularity of large language models and their usefulness in many NLP domains~\citep{devlin-etal-2019-bert}, the representation of temporal knowledge in those models so far remains an open challenge.

The increased interest in temporal adaptation (i.e. scenarios in which the training and test datasets are drawn from different periods of time) has led to the curation of a number of datasets such as NYT Annotated Corpus~\citep{nyt-2008} and Amazon Reviews~\citep{amazon-review} that have been the focus of most of the recent work in this area. 
However, these benchmark datasets are curated in such a way that they can only capture temporal change of language over long periods of time (from years to decades), giving access to a large amount of data. 
In the contrary, on social media, language changes can happen rapidly~\citep{kulkarni2015, eisenstein2013}. 
Word usage and topics can even change over the span of a single day~\citep{golder2011}, especially during very dynamic scenarios like crisis or disastrous events~\citep{reynolds2005, del-tredici-etal-2019-short}. 
We denote these phenomena induced by linguistic and semantic changes over time as \textit{temporal drift}.

Accounting for temporal drift is critical in crisis situations in which information patterns can vary greatly between the phases of emergency management for crisis. 
For this purpose, we study short text classification in crisis situations.
Given the time-critical nature of crisis scenarios, gathering annotations is too time-consuming and transfer learning is challenging due to the innate differences among the type of events (hurricane vs. earthquake) and the respective information needs.
Thus, we offer a study investigating the impact of temporal drift on crisis datasets spanning shorter time periods (days/weeks), as well as datasets with relatively few samples (ranging from $\sim$1k to 22k). 
\begin{figure*}
    \centering
    \includegraphics[width=\textwidth]{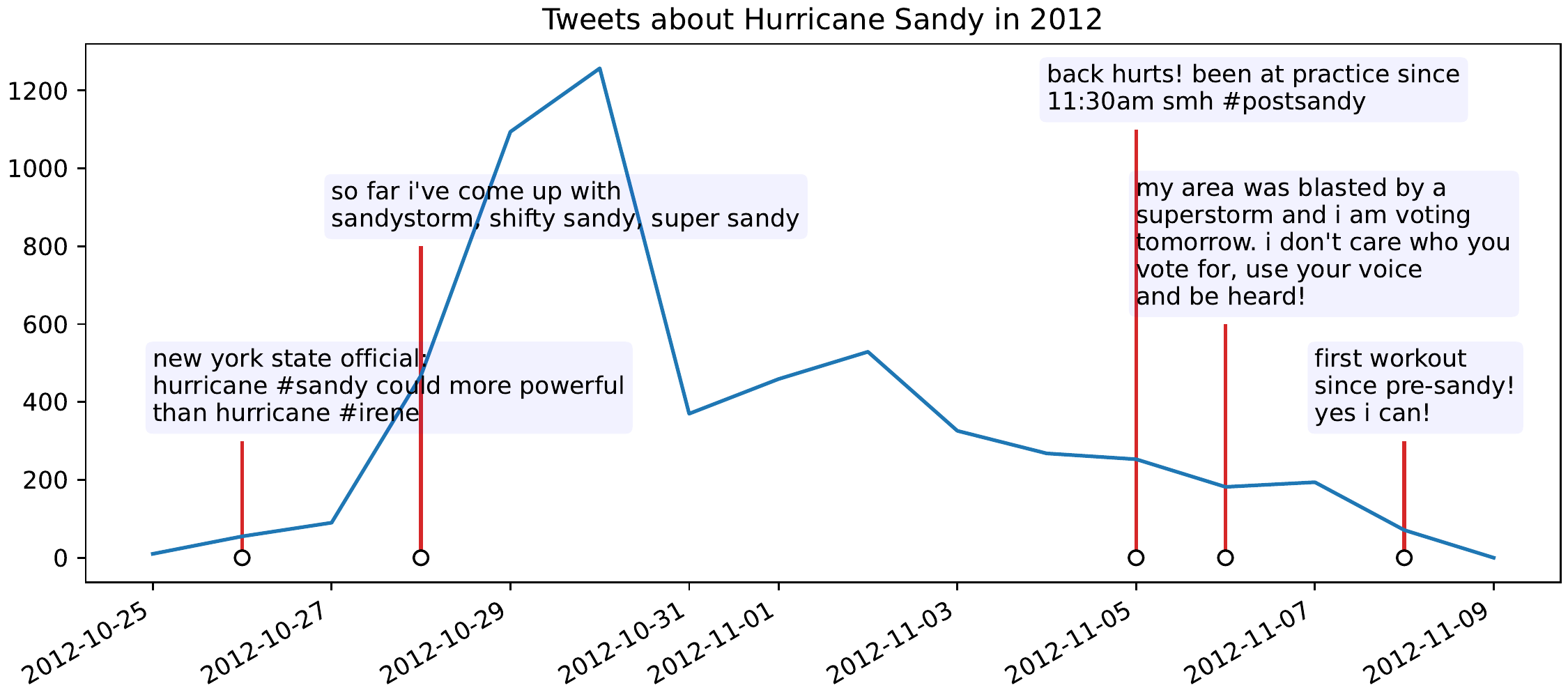}
    \caption{The blue line indicates the frequency of tweets during the hurricane Sandy \cite{stowe2018}. The displayed tweets demonstrate challenging linguistic phenomena for a text classification model, e.g. semantic shifts (\textit{\#irene} as reference to a hurricane rather than a person) or neologisms (\textit{pre-sandy}).}
    \label{fig:timeline}
\end{figure*}

Assessing rapid temporal drift is a challenging problem due to different linguistic phenomena which often require extensive knowledge about the temporal structure of the context. 
In Figure \ref{fig:timeline}, we provide examples from \citet{stowe2018} dataset, which were collected from the 2012 New York City landfall of Hurricane Sandy.

Existing approaches like continual learning \cite{gururangan-etal-2020-dont, loureiro2022timelms} or learning time-specific models \cite{agarwal2021temporal} cannot be applied to this scenario as access to a large set of unlabeled data from the temporal target distribution is missing.
Unlike existing approaches, which react to incoming annotated data to update their models, we use temporal metadata as a training signal such that the existing contextualized representations are adapted temporally.
More specifically, we are the first to apply projection methods~\citep{Wang_Zhang_Feng_Chen_2014} and domain adaptation approaches~\citep{ganin2016domain, bamler2017dynamic} to learn time-aware contextualized embeddings.
Our results highlight the challenges of integrating temporal information into contextualized embeddings with improvements being dependent on factors like dataset size - and thereby emphasizing that temporal adaptation remains a challenge in scenarios where we do not have access to large unlabeled data. 

In summary, we make the following contributions:
\begin{enumerate}
    \item We investigate temporal drift during crisis events and its adversarial effect on task performance. To the best of our knowledge, this is the first study of temporal effects on text classification performance in crisis scenarios, when temporal drift is rapid and access to data is scarce.
    \item We investigate the role of the domain of data in temporal drift and propose a simple metric to quantify the impact of temporal degradation on task performance.
    \item   We propose methods that adapt future data to known models, improving performance with no additional labeled data.
    \item Through experiments on a multitude of diverse text classification datasets collected during crisis events, we analyse the effectiveness of our proposed methods over strong baselines.
    
\end{enumerate}

\section{Related Work}
\label{sec:rw}

Analyzing semantic change of text over time has been of great interest since the pioneering work by ~\citet{hamilton-etal-2016-diachronic} and others~\citep{kutuzov-etal-2018-diachronic, 10.1145/3178876.3185999, martinc-etal-2020-leveraging, gonen-etal-2020-simple}.
However, its influence on downstream task performance has only recently gained attention.
Most importantly, the advent of contextualized word embeddings and large pretrained language models has led researchers to re-evaluate the role of temporality in language modeling~\citep{jawahar-seddah-2019-contextualized, lazaridou2021mind, hofmann-etal-2021-dynamic, kulkarni-etal-2021-lmsoc-approach} and text classification~\citep{bjerva2019, app10124180, rottger-pierrehumbert-2021-temporal-adaptation, agarwal2021temporal}. 

The performance degradation due to temporal factors has been confirmed in several studies and across multiple domains.
\citet{jaidka-etal-2018-diachronic} analyzed the temporal performance degradation of age and gender classification models based on user's social media posts. 
Based on features derived from Latent Dirichlet Allocation and word embeddings, they find that models perform best if test and training data come from the same time span. 
\citet{app10124180} investigated temporal effects on Hate Speech detection in Italian social media over the period of five months. 
Their results suggest that models trained on data temporally closer to the test data perform better with transformer based models.
\citet{tempowic_coling22} studied semantic shifts in social media and proposed a dataset annotated with words that have undergone a semantic shift over the past two years. \citet{loureiro2022timelms} focus on Twitter as text domain and contribute pretrained language models which have been further trained on time-specific data from Twitter.

\citet{bjerva2019} propose to use sequential subspace alignment (SSA) to adapt contextualized word embeddings for language change over time.
Their results suggest that SSA applied on past data is able to outperform baselines which have access to data from all time-steps.
\citet{rottger-pierrehumbert-2021-temporal-adaptation} compared time-agnostic domain adaptation with temporal domain adaptation which considers the temporal order of the data. 
They found that, while temporal adaptation clearly outperforms domain adaptation in language modeling, this does not necessarily translate onto downstream classification performance due to updated tokens not being relevant for the task.
\citet{agarwal2021temporal} found the temporal model performance deterioration to be less significant when using language representations which have been pretrained on temporally closer data.

Finally, \citet{luu2021time} have made the effort of conducting a large-scale study of temporal misalignment, the generalized scenario where training and evaluation data are drawn from different periods of time.
Across multiple NLP classification tasks and domains they identify performance degradation with varying degrees but with social media and news being the most affected domains.

We contribute to the existing line of work by quantifying the temporal effects on downstream task performance over short time periods (days and weeks) during crisis events.
In such a scenario and in contrast to previous work, we do not assume access to large corpora of unlabeled data for temporal adaptation via continuous pretraining.
Our proposed approaches temporally adapt pretrained contextualized embeddings to learn time-aware embeddings and we evaluate their effects on downstream classification tasks.

\section{Methods Overview}
\label{sec:scenario}

\citet{luu2021time} describe three distinct stages of a typical NLP system which consist of a pretraining stage, a domain (or temporal) adaptation stage and a fine-tuning stage.
Separating the adaptation and fine-tuning stages makes the implicit assumption that there is access to unlabeled data from the (temporal) target distribution which has been proven to be beneficial for temporal adaptation~\citep{luu2021time}.
In contrast, we are looking at the dynamic setting during crisis events.
Temporal alignment through continuous pre-training is not feasible due to the lack of unlabeled data and time constraints imposed by the application scenario (e.g. crisis monitoring).
The latter also limits the feasibility of an online learning setup which requires new annotations in a continuous stream.
Finally, transfer learning is difficult due to inherent differences in information needs (i.e. the type of labels) and domains (e.g. hurricane vs. earthquake).

Therefore, in this section we adapt and evaluate methods which are specifically designed for combining temporal adaptation and fine-tuning. 
Their training procedures are adapted to incorporate temporal information about the data along with the textual input.
We describe each approach in the following:
\label{ssec:models}

\subsubsection*{Adapted Language Modelling (ALM)}
\label{sssec:alm}

Similar to previous work (see Section \ref{sec:rw}), we explore temporal adaptation via pretraining but use only the available training data.
We therefore continue with the language modeling objective of our respective pretrained language model on the training data and use the resulting fine-tuned model (\textbf{FT}) for downstream task training.
Following \citet{dhingra2021time}, we investigate a variation for temporal modelling (\textbf{TM}) by concatenating time as textual information to the input to encourage the language model to learn temporally relevant features during pretraining.

\subsubsection*{DCWE: Dynamic Contextualized Word Embeddings}

\citet{hofmann-etal-2021-dynamic} introduced a principled way to impart extra-linguistic knowledge into contextualized word embeddings by involving a prior distribution. 
This enables us to integrate temporal information into the embeddings during training.\footnote{Other extra-linguistic information like social context can also be integrated.}
More specifically, for each temporal snapshot (e.g. days, months, years, etc.) present in the training data, an additional set of parameters is learned which acts as a temporal offset added to the original word embeddings.
This way the model is able to maintain the semantic meaning of a word embedded in its temporal context.
We adapt this idea to our setting by introducing additional parameters for shifting the pre-trained contextualized embeddings.
Given a sequence of words/tokens $W = [w_1, w_2, ..., w_n]$ and their corresponding pre-trained embeddings $H = [h_1, h_2, ..., h_n]$. 
To account for the temporal effect on the word meanings, we model word embeddings as a function of temporal context $t$ associated to $W$. 

\begin{equation}
	h^{*}_{i} = f(h_i, t)
\end{equation}

Since meanings of most of the words in the vocabulary are temporally stable, we can place a Normal prior on $h^{*}_{i}$.

\begin{equation}
	h^{*}_{i} \sim \mathcal{N}(h_i, \lambda^{-1}I)
\end{equation}
 
Hence, we write as $h^{*}_i = h_i + d_i$, where the offset $d_i$ is normally distributed as $d_i \sim \mathcal{N}(0, \lambda^{-1}I)$. 
However, pre-trained LMs make this temporal adaptation easily applicable to any task by adding only a regularization term $L_{temporal}$ on top of the task specific loss $L_{task}$. 

\begin{equation}
	L_{temporal} = \frac{\lambda}{n}\sum_{i=1}^{n}(||d_i||_{2}^{2} + K||d_i - d_{i-1}||_{2}^{2})
\end{equation}

For training the model, the overall loss $L = L_{task} + L_{temporal}$ is minimized. 
Similarly to \citet{hofmann-etal-2021-dynamic}, we use $K=10^{3}$ from \citet{bamler2017dynamic}, to enforce that $h^{*}_{i}$ s change smoothly over time.

\subsubsection*{LMSOC: Socio-temporally Sensitive Language Modeling}
Similar to DCWE, \citet{kulkarni-etal-2021-lmsoc-approach} propose a method to learn extra-linguistic context using graph representation learning algorithms and then primes with language models to generate language representations grounded in a socio-temporal context. 
We model the temporal order information as a linear chain graph and adapt this method to our setting by appending temporal graph embeddings to the initial layers of the pre-trained language model. 
During fine-tuning of the language model, the graph embeddings are kept frozen to inductively yield temporally-aware embeddings. 

\subsection*{TAPH: Time Aware Projection on Hyperplanes}

Time adds an additional context or dimension to the knowledge making temporal scoping an imperative part while deriving context embeddings.
Therefore, we model temporal information as a hyperplane and define a projection operation \citep{Wang_Zhang_Feng_Chen_2014} on it. 
To build a time-invariant classification model, we project the sentence-embedding \citep{reimers-2019-sentence-bert} of each text on a hyperplane to obtain a time-aware sentence embedding. 
We describe the method in more detail.
 
Let $X = [x_1, x_2, ..., x_n]$ be a given sequence of words and $H$ be its sentence embedding. 
Since the temporal span of our data is short, we assume that the temporal hyperplane $w_t$ represents the time frame of the training data.\footnote{For longer time spans it is possible to divide the training data into multiple static bins.}
We derive time-aware sentence embeddings $H_t$ using our defined projection operation as follows:

\begin{equation} \label{eqn:taph}
	H_t = H - w_{t}^{\intercal}Hw_{t}
\end{equation}

While training the model, we learn the hyperplane representation $w_t$ in addition to fine-tuning the pre-trained embeddings in an end-to-end fashion. 
However during inference, we assume that we could `teleport' the data to the past by projecting their sentence embeddings on the hyperplane $w_t$ in order to revert their temporal changes. 
We then use these embeddings in the downstream tasks. 

\subsubsection*{TDA: Temporal Domain Adaptation}

Temporal Adaptation can also be interpreted as a variant of domain adaptation with the difference that the language change happens within the same domain, e.g. induced by external events or the general dynamic characteristics of the source infrastructure (e.g. social media platforms or news outlets).
We adapt a widely used domain adaptation method \citep{ramponi-plank-2020-neural} to our setting.
We learn time-aware word representations by adding an additional classification layer during training to predict the time of each text and apply the Gradient Reversal method~\citep{ganin2016domain}.
In this way, the input does not change during the forward pass but this additional layer affects the model parameters during back-propagation of error by an additional penalizing factor.\footnote{During the back propagation its corresponding gradients are multiplied with a negative scalar (hyperparameter $\lambda$)} 
This acts as an adversarial training objective forcing the model to adapt to the temporal structure of the data.

\section{Experimental Setup}
\label{sec:exp_setup}

\subsection{Data}
\label{ssec:data}

We identify a collection of social media data during crisis with observable temporal phases (pre-, acute- and post-crisis), rapid change in language and a natural change in distribution over time - enabling us to evaluate how well temporally adapted models generalize over time. 
We use three datasets sampled from Twitter: \sandy{}, \clex{}, and \humaid{}. 
We provide an overview here and refer to the Appendix~\ref{sec:appendixa} for dataset details.

\paragraph{\sandy{}}
The dataset from \citet{stowe2018} collected during hurricane Sandy in 2012 contains approximately 22,000 tweets spanning 17 days centered on landfall in New York City, annotated for binary relevance to the storm and its effects.\footnote{\url{https://github.com/Project-EPIC/chime-annotation}}
The tweets were collected by first identifying users impacted by the event, then retroactively pulling their data from before, during, and after the event. 
As opposed to keyword collection, this provides a relatively broad collection of both relevant and non-relevant tweets and a more complete dataset for evaluating temporal drift, as each tweet doesn't necessarily contain the same keyword(s).

\paragraph{\clex{}}
The CrisisLex T26 (\clex{}) dataset \cite{olteanu2015} includes labeled tweets for 26 different crisis events, labeled by informativeness into four different categories\footnote{\url{http://www.crisislex.org/data-collections.html\#CrisisLex}}: (1) related to the crisis and informative, (2) related to the crisis but not informative, (3) not related to the crisis, and (4) not applicable category. 
This collection reflects a wide variety of events covering natural and human-created emergencies, with the added difficulty that the individual datasets are relatively small, with each event containing only approximately 1,000 tweets. 

\paragraph{\humaid{}}
The \humaid{} dataset~\citep{alam2020} is similar to \clex{}, containing data about 19 different events with dataset sizes ranging from 575 to 9467 tweets. 
They are annotated with 11 different classes designed to capture fine-grained information related to disaster events. 

\subsection{Data Splits}

\begin{figure}
    \centering
    \includegraphics[width=0.48\textwidth]{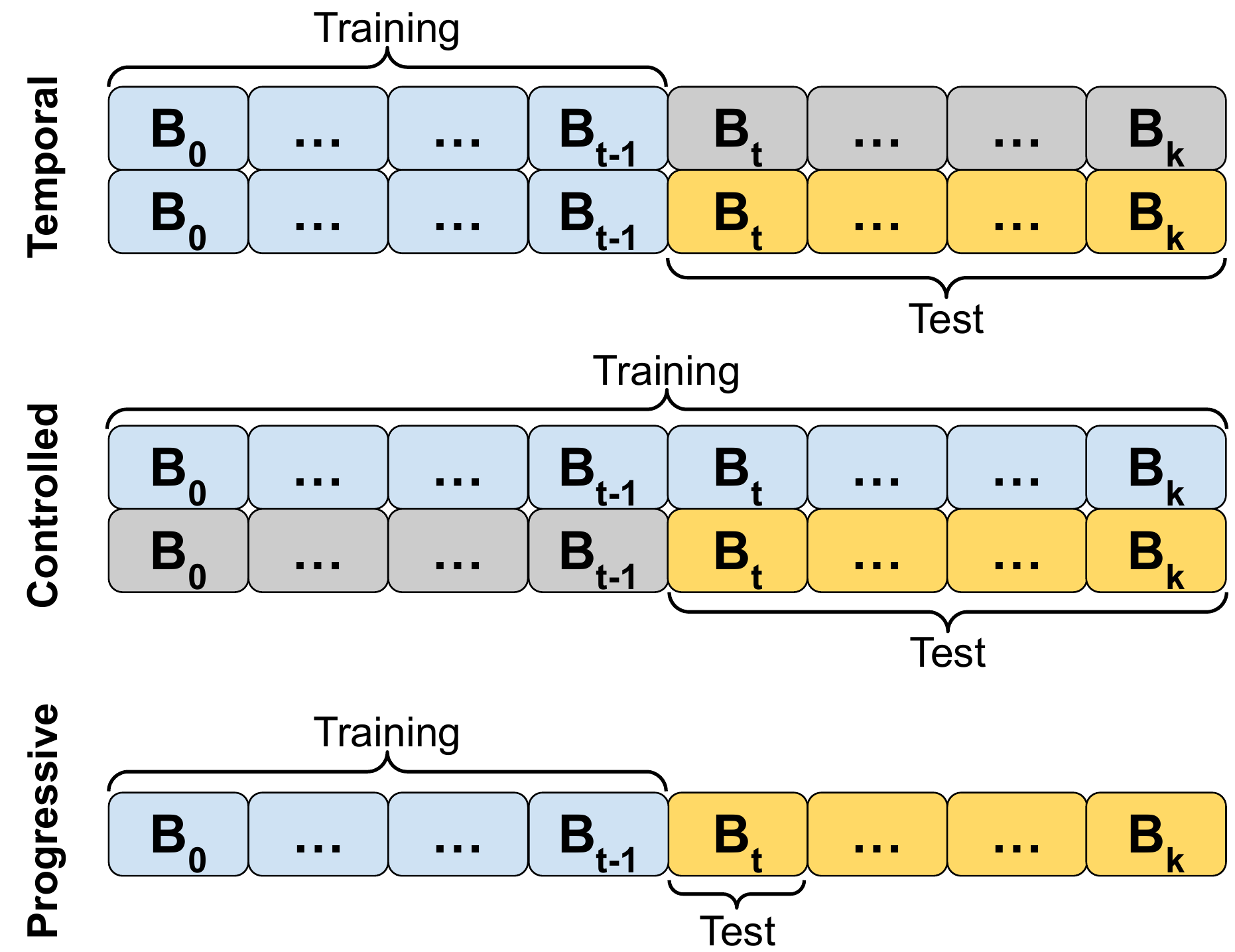}
    \caption{Overview of the data splits used in our experiments. Bins in blue are used during training, bins in yellow for testing, grey bins are not used. The \prog{} setting comprises multiple experiments with increasing training data size and a single test data bin moving forward temporally.}
    \label{fig:datasplits}
\end{figure}

We follow previous work~\citep{lazaridou2021mind, agarwal2021temporal} and create time-based data splits to assess the temporal performance degradation. 
Specifically, we use three variants of dataset splits: \cont{}, \temp{} and \prog{}. We illustrate this in Figure~\ref{fig:datasplits}.

\paragraph{\temp{} Setup} 
First, we split the entire data into two halves which cover equally-sized time periods.
We call these first temporal half and the second temporal half, respectively. 
In the \temp{} setting, we use all the data from the first temporal half as the training data and a test set which is comprised of a randomly sampled $50\%$ of data from the second temporal half of a dataset.
This evaluates the model's temporal generalization capabilities on test data from a temporally distant distribution than the training data.

\paragraph{\cont{} Setup}
To assess whether \temp{} setup constrains model's generalization capabilities, we compare its performance with a \cont{} setup. 
Here, we evenly spread the training data over time frames, exposing the model to the full knowledge of all time. 
In this setting, the training data comprises of $50\%$ of instances from the first temporal half, along with $50\%$ instances from the second temporal half, matching the total training data from the \temp{} setup.
We use the same test set as in \temp{} setup while ensuring that there is no overlap between the train and test split from the second temporal half.

Under the assumption that a temporal gap between training and target distribution leads to performance decay, we expect that the \cont{} setup will yield better scores, as the model has access to training instances from the same temporal distribution as the test data.

\paragraph{\prog{} Setup}
As described previously, semantic changes are likely to occur in short time spans within crisis-related data streams.
Therefore, to investigate a more fine-grained analysis of temporal performance decay, we simulate a scenario in which an event is progressing, we have access to all the previous data, and need to take decisions about the incoming data.
In this setup, we split the entire dataset into ten temporally ordered bins with even samples. 
Then, for each test bin $B_{t}$, we use all preceding bins $B_0$ to $B_{t-2}$ for training.
To identify the best performing model across all training epochs, we use bin $B_{t-1}$ for development.

\subsection{Baseline}

For a consistent performance comparison, all proposed models use \texttt{bert-base-cased} as their underlying backbone model for deriving pre-trained embeddings.

For the \textbf{FT} setup (see Section~\ref{sssec:alm}), we use the available training data for each dataset to run masked language modelling for three epochs to adapt the model to the data.
We then fine-tune for the downstream task on the relevant training data using the updated pre-trained model.
This will indicate whether the domain is the issue, or whether there is additional temporal effects. 
In the temporal modeling setup (\textbf{TM}) setup, we follow \citet{dhingra2021time} and prepend the textual representation of the timestamp for each tweet to the tweet text, then train an additional three epochs of masked language modelling.
We then fine-tune for the downstream task on the relevant training data.

Finally, we apply another baseline where we use the timestamp text as second input to the model during supervised training, separated via a special token (i.e. \texttt{[SEP]} for BERT).
We refer to this baseline as \textbf{SEP}.

\subsection{Hyperparameters and Infrastructure}

For a fair comparison, we run all experiments using the same hyperparameters and data splits. 
We use a learning rate of $1e-4$, batch size of 64, weight decay of $1e-3$ and no warmup due to the limited amount of training data.
We use Adam~\cite{loshchilov2018decoupled} as optimization algorithm and train for three epochs.
Based on the performance on the development split, we load the best performing model at the end of the training procedure.

We repeat each experiment using five different seeds and take the most frequent prediction across all runs as the final prediction by a model.
All models are implemented in Python 3.6 using PyTorch 1.10.2~\citep{pytorch2019} and the HuggingFace~\citep{wolf-etal-2020-transformers} framework (4.18) as model backend.
We used a computation cluster containing a mixture of NVIDIA Tesla P100 (16GB), NVIDIA A100 (40GB) and NVIDIA V100 (32GB) GPUs.

\subsection{Evaluation} 
\label{subsec:eval}

We report binary-F1 Score for \sandy{} and macro-F1 score for multi-class classification task on \clex{} and \humaid{} datasets.
The comparison of the \cont{} and \temp{} setting serves two purposes; first, to quantify the degradation of model performance due to temporal drift and second, to estimate the temporal adaptation ability for our approaches.
We expect that models considering temporal information should experience less performance degradation between these two settings compared to the baseline model. 

Additionally, we evaluate the mean model performance in the \prog{} setting for a more fine-grained analysis of temporal degradation.\\

\noindent{\bf Temporal Rigidity:} While analyzing the effects of temporal drift on model performance, it is necessary to quantify the degradation of model performance due to this phenomenon. 
We quantify the temporal adaptability of a model using a metric called \textit{Temporal Rigidity} (TR) score, that summarizes the performance deterioration of a model from aligned to misaligned test data.  
Higher values of TR imply that the model is not able to adapt itself temporally.

We denote $f_{M}(B_i, B_j)$ as the F1 performance score of a model $M$ when trained using data sampled from bin $B_i$ and evaluated using data sampled from bin $B_j$. We define TR as:

\begin{equation} \label{eqn:tr_score}
	TR = \frac{1}{N}\sum_{i \neq j}{\frac{\lvert f_{M}(B_i, B_j) - f_{M}(B_i, B_i) \rvert}{\lvert i - j \rvert}}
\end{equation}

In Eqn.\ref{eqn:tr_score} the normalization factor is given as $N=\lvert \{(i, j): i\neq j\} \rvert$. 
Unlike \citet{luu2021time}, who do not take temporal proximity of bins into account.
We use $\frac{1}{\lvert i-j \rvert}$ as the penalizing factor for the model when training and test bins are temporally close but the performance degradation is significant.

\noindent{\bf Crisis Phases:} Additionally, we utilize the well-known temporal structures of the crisis events~\citep{reynolds2005,yang2013} to analyze model performance. 
The temporal structure of the \sandy{} dataset is annotated using {\it pre-}, {\it acute-} and {\it post-crisis} labels. 
For each model we cluster the time-aware embeddings using K-Means algorithm (k=3) and report the Normalized Mutual Information score (NMI). 
NMI gives the correlation between the time-aware embeddings and the temporal structure of the underlying data. 

\section{Results and Analysis} \label{sec:result}

In this section, we attempt to answer the following questions:

\begin{enumerate}[start=1,label={Q\arabic*.}]
    \item To what degree is temporal performance degradation present in short-term Twitter data during crisis events? (Section \ref{ssec:temporal_degradation})
    \item Does temporal adaptation improve model performance? (Section \ref{subsec:performance_comp})
    \item Does the domain of the data play a role in temporal drift? (Section \ref{subsec:data_domain})
    \item How do the proposed models perform when trained continually? (Section \ref{subsec:cont_learn})
\end{enumerate}

\subsection{Temporal Performance Degradation}
\label{ssec:temporal_degradation}

\begin{table}[!ht]
	\begin{center}
		\begin{tabular}{l c c c}
			\toprule
			
			Data & \sandy{} & \clex{} & \humaid{} \\
			
			\midrule
			
			\begin{small} \cont{} - \temp{} \end{small}& 6.52 & 4.37 & 4.10 \\
			
			\bottomrule
			
		\end{tabular}
	\caption{{\bf Temporal Performance Degradation:} Averaged F1 performance difference of the \cont{} to \temp{} setting for the BERT baseline model. Overall results show that contextualized language models fail to adapt temporally. Refer Section \ref{ssec:temporal_degradation} for details.}
	\label{tab:diff}
	\end{center}
\end{table} 

In order to estimate the degree of temporal performance degradation in the crisis scenario, we compare the classification performance of the baseline model in the \cont{} and \temp{} setting.
Table~\ref{tab:diff} provides the averaged performance difference for all datasets.
Given that we only change the temporal distribution of the training data, the effect is substantial with a difference in F1 up to 6.52 points for the \sandy{} dataset and slightly less pronounced on the \clex{} (4.37) and \humaid{} (4.10) dataset collections.
Therefore, we conclude that, even in short-term scenarios like crisis events on Twitter, temporal distribution of the training data influences the classification performance. 

\subsection{Performance Comparison}
\label{subsec:performance_comp}


\begin{table}[h!]
    \begin{center}
        \begin{tabular}{l c c c}
            \toprule
            
            \multirow{2}{*}{Method} & \multicolumn{3}{c}{\sandy} \\
            
            \cmidrule(lr){2-4}
            
            & \textsc{cont} & \textsc{temp} & \textsc{diff}\\
            
            \toprule
            
            BERT & 87.70 & 81.18 & 6.52 \\
            
            BERT+TM & 82.55 & 70.48 & 12.07 \\
            
            BERT+SEP & 87.79 & 79.65 & 8.14 \\
            
            \midrule
            
            BERT+LMSOC & 73.78 & 67.24 & 6.54 \\
            
            BERT+DCWE & 86.92 & 79.95 & 6.97 \\
			
			BERT+TAPH & 87.40 & 82.02 & 5.38 \\
			
			BERT+TDA & 87.10 & \bf{82.53} & \bf{4.57} \\
            
            \midrule
			
			BERT\textsubscript{FT} & 86.96 & 81.84 & 5.12 \\
			\midrule
			
			BERT\textsubscript{FT}+LMSOC & 74.89 & 67.90 & 6.99 \\
			
			BERT\textsubscript{FT}+DCWE & 86.85 & 79.53 & 7.32 \\
					
			BERT\textsubscript{FT}+TAPH & 87.12 & 82.60 & 4.52 \\
			
			BERT\textsubscript{FT}+TDA & 86.71 & \bf{83.43} & \bf{3.28} \\
			
			\bottomrule
             
        \end{tabular}
        \caption{\label{tab:eval_sandy} {\bf Temporal Adaptation Evaluation on \sandy{}:} Text classification performance measured in binary F1. Overall, TDA outperforms other approaches in \temp~ setting, with and without temporal adaptation (FT). Refer Section \ref{subsec:performance_comp} and \ref{subsec:data_domain} for details.}
    \end{center}
\end{table}

\begin{table}[!ht]
	\begin{center}
		\begin{tabular}{l c c c}
			\toprule
			
			Method & \clex{} & \humaid{} \\
			
			\midrule
			
            BERT+TM & 4 / 26 & 3 / 19 \\
            
            BERT+SEP & 5 / 26 & 3 / 19 \\
            
            \midrule
            
            BERT+DCWE & 0 / 26 & 1 / 19 \\
			
			BERT+TAPH & 6 / 26 & 0 / 19 \\
			
			BERT+TDA & \textbf{10 / 26} & \textbf{4 / 19} \\
            
            \midrule
			
			BERT\textsubscript{FT}+DCWE & 0 / 26 & 0 / 19 \\
					
			BERT\textsubscript{FT}+TAPH & 5 / 26 & 0 / 19 \\
			
			BERT\textsubscript{FT}+TDA & 8 / 26 & 0 / 19 \\

			\bottomrule
			
		\end{tabular}
	\caption{{\bf Performance Comparison on \clex{} and \humaid{}: }The number of datasets for which the specific temporal adaptation method outperforms its baseline counterpart in the \temp{} setting. Refer Section \ref{subsec:performance_comp} and \ref{subsec:data_domain} for details.}
	\label{tab:clex_humaid}
	\end{center}
\end{table} 

We summarize the results on \sandy{} in Table \ref{tab:eval_sandy}. 
Overall we find that TDA outperforms all other methods in \temp{} setting. 
We obtain around $1.6\%$ absolute increase over the baseline. 
We also observe that the difference between model performance in \cont{} and \temp{} setting (\textsc{diff}) is lowest for TDA ($30.8\%$ lower than the baseline) indicating the higher robustness of the model. 
TAPH achieves $1\%$ absolute improvement in performance over the baseline in \temp{} setting (\textsc{diff} is lower by $16.9\%$).

The \clex{} and \humaid{} datasets contain data for a multitude of events.
Therefore, we aggregate model performances in Table~\ref{tab:clex_humaid} and provide detailed results per event in the Appendix~\ref{app:results}.
We see that model performance varies greatly between the \sandy{} dataset and the others. 
This is due to two main reasons: (i) {\bf Data Size: }Most of the event datasets in \clex{} and \humaid{} are very small, the temporal adaptation methods do not get enough training data to learn the parameters involved in temporal reasoning. 
To support our argument, we observe, in ``Boston Bombings (2013)'' dataset of \clex{}, which contains 81,172 annotated tweets, TDA outperforms the baseline by an absolute increases of $6.17\%$ and TAPH comes second with an absolute performance improvement of $2.9\%$ under \temp{} setting, a performance pattern which is similar to \sandy{} dataset.
(ii) {\bf Data Quality: } Unlike \sandy{}, \clex{} and \humaid{} have been collected using keyword-based search. 
This data collection technique has two main drawbacks: firstly, it restricts the data size and secondly, harms the completeness of the dataset collecting tweets that contain same keywords. 
All the improvements we report are statistically significant ($p < 0.05$, using McNemar's Test).\newline

\noindent {\bf Learning from Temporal Information:} To understand the cause of the performance improvement of the models, we utilize the annotated temporal structure of the \sandy{} dataset. In Table \ref{tab:temp_learning} we report two additional metrics: TR Score and NMI, in \temp{} setting. 
Compared to the baseline, TDA is lowest ($15.74\%$ decrease) which suggests that TDA performs most robustly over time across all models. 
TAPH comes in second with a $9.26\%$ decrease in TR Score from the baseline.
NMI scores show similar patterns, with TDA achieving the highest score. 
We conclude that TDA learns the most meaningful time-aware embeddings.

\begin{table}[!ht]
	\begin{center}
		\begin{tabular}{l c c}
			\toprule
			
			\multirow{2}{*}{Method} & \multicolumn{2}{c}{\sandy{}} \\
			
			\cmidrule(lr){2-3}
			
			& TR & NMI \\
			
			\toprule
			
			BERT & 0.108 & 0.051\\
			
			BERT+TM & 0.130 & 0.050 \\
			
			\midrule
			
			BERT+DCWE & 0.111 & 0.105\\
			
			BERT+TAPH & 0.098 & 0.185\\
			
			BERT+TDA & {\bf 0.091} & {\bf 0.194}\\
			
			\bottomrule
			
		\end{tabular}
	\caption{\label{tab:temp_learning}{\bf Temporal Information Learning:} Comparison of methods on TR (lower is better) and NMI scores (higher is better). Refer section \ref{subsec:performance_comp} for details.}
	\end{center}
\end{table}

\subsection{Effect of Domain of Data}
\label{subsec:data_domain}

To understand whether the data domain is the main issue behind performance degradation or temporal effects indeed play a significant role, we perform additional experiments. 
We fine-tune the initial \texttt{bert-base-cased} embeddings for an additional three epochs with Masked Language Modeling Task (MLM) on the training data, before applying the Temporal Adaptation methods. 
We report the results for \sandy{} dataset in Table \ref{tab:eval_sandy}. 
For all models, there remains a substantial performance difference between the \cont{} and \temp{} settings which demonstrates the influence of temporal drift on performance.
Similar to previous work~\citep{agarwal2021temporal}, we observe that additional pre-training improves performance for most of the models.
Still, TDA outperforms the baseline and TAPH comes in second. 

\subsection{Effect of Continual Learning:}
\label{subsec:cont_learn}

\begin{table}[!ht]
    \centering
    \begin{tabular}{l c}
    \toprule
    Method & \sandy{} \\
    \midrule
    BERT & 68.67 \\
    
    BERT+TM & 60.13\\
    
    \midrule
    
    BERT+DCWE & 67.39 \\
    
    BERT+TAPH & 69.13 \\
    
    BERT+TDA & {\bf 69.50} \\
    
    \bottomrule
    
    \end{tabular}
    \caption{\label{tab:prog-all}{\bf Continual Learning Effects: } Average model performance across all bins in \prog{} setting, in terms of F1 Score. Refer section \ref{subsec:cont_learn} for details.}
\end{table}

Continual Learning requires continuous annotation of incoming data, which is not feasible during crisis events. 
However, for the analytical completeness of this paper, we simulate continual learning in the \prog{} setting to show the effectiveness of our proposed methods. 
In this setting, initially the models get access to very small amount of data to learn from, which affects model performance. 
Performance improves as the size of training data increases gradually. 
In Table~\ref{tab:prog-all} we report the model performance averaged over all the bins. 
The results show that TDA outperforms and improves the BERT baseline by $1.2\%$.

\section{Discussion}
\label{sec:discussion}

\begin{figure}
    \centering
    \includegraphics[scale=0.5]{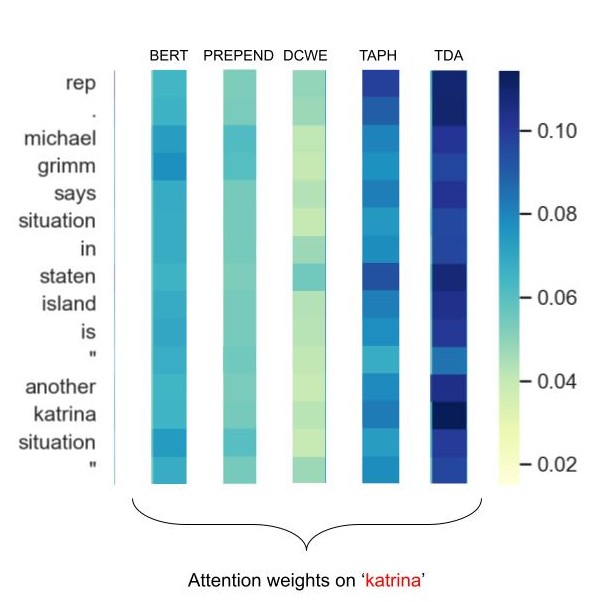}
    \caption{Representative example shows that in comparison with other models TDA correctly puts maximum attention weight on the word \textcolor{red}{katrina} (another storm) in the temporal context of the hurricane while computing the contextual embeddings. Refer Section \ref{sec:discussion} for details.}
    \label{fig:attention_eg}
\end{figure}

Adapting temporally by training on timestamp patterns as text prepended to the input (BERT+TM) underperforms in all experiments. 
We argue that the added information affects all tokens equally via the self-attention mechanism  although only some tokens will experience a semantic shift relevant for text classification in the crisis scenario.

Similarly, the LMSOC and DCWE adaptation approaches cannot outperform the baseline without any temporal adaptation.
The additional parameters for computing the temporal offset are not well-tuned for predicting temporal distributions which have not been observed during training.

Figure \ref{fig:attention_eg} shows that TDA correctly learns to put maximum attention weight on the word \emph{Katrina} (i.e. reference to a previous hurricane) in the temporal context of hurricane. 
We provide representative examples of tweets in Appendix \ref{sec:appendixa} that all other models but TDA fail to classify correctly.
Forcing the model to learn time-invariant embeddings during training using an adversarial signal leads to TDA performing better over all other approaches. 
Although, TAPH does not fall far behind, it approximates temporal information to create time-static bins. 
The discrete approximation of temporal information is the main reason behind its performance drop. 

\section{Conclusion}

The usage of natural language inevitably changes over time which influences performance of text classification models applied on data from different temporal distributions.
We show that this effect is also prevalent for rapid temporal drift using social media during crisis events as an example.
With the rise of pretrained contextualized embeddings, a dominant approach is to continue language modeling on data temporally closer to the target distribution.
However, during crisis events such data is not available and annotated data is often scarce.

We investigate approaches which work without any additional data besides the input text and its temporal metadata.
Our results show that under ideal conditions, i.e. high data quality and sufficient annotated instances, they outperform strong baselines.
However, most crucially, our work highlights a critical gap of temporal adaptation for rapid temporal drift, namely if unlabeled data for alignment is missing and annotated data is scarce.
Our work opens the door for future research on methods which do not rely on pretraining in unlabeled target domain data.
In this sense, crisis data provides an interesting use case for evaluation.
We release all our code and models, fostering future work in this area. 

\section*{Limitations}

While existing approaches account for temporal change of language over long periods of time, in social media this change can happen over the span of a single day during dynamic scenarios like crisis or disastrous events. 
In this work we study rapid temporal drift prevalently observed in social media during a crisis. 
We observe that often data from social media are collected using keyword based search and data sampling techniques, where data containing same set of keywords are collected. 
Since data collected using such techniques are both limited by size and vocabulary, as well by the issues inherent in keyword collection, the datasets naturally affect the performance of the methods described in the paper. 
Moreover, there exists differences among the types of crisis events (hurricane vs. earthquake) and their respective information needs. 
Hence, it is difficult to find a solution that works in all scenarios. 
Additionally, we highlight that evaluation of all the models was done on datasets annotated in presence of a crisis and that may not exactly reflect their performance in a real-world setting without annotated data, especially when differences among the types of crises are relevant. 
In a nutshell, we observe that during real-world crises, pre-trained language models turn out to be a good solution when access to unlabeled data is scarce and sufficient annotated data is unavailable.




\section*{Acknowledgements}
We thank Ilia Kuznetsov, Jan Buchmann, Luke Bates and the anonymous reviewers for their valuable feedback and Firoj Alam for providing us full access to the HumAID dataset.
This work has been funded by the German Research Foundation (DFG) as part of the Research Training Group KRITIS No. GRK 2222. It has further been funded by the project “Open Argument Mining” (GU 798/25-1), associated with the Priority Program “Robust Argumentation Machines (RATIO)” (SPP-1999), and by the LOEWE initiative (Hesse, Germany) within the emergenCITY center.

\bibliography{anthology,emnlp2022}

\begin{thebibliography}{41}
\expandafter\ifx\csname natexlab\endcsname\relax\def\natexlab#1{#1}\fi

\bibitem[{Agarwal and Nenkova(2022)}]{agarwal2021temporal}
Oshin Agarwal and Ani Nenkova. 2022.
\newblock \href {https://doi.org/10.1162/tacl_a_00497} {Temporal effects on
  pre-trained models for language processing tasks}.
\newblock \emph{Transactions of the Association for Computational Linguistics},
  10:904--921.

\bibitem[{Alam et~al.(2021)Alam, Qazi, Imran, and Ofli}]{alam2020}
Firoj Alam, Umair Qazi, Muhammad Imran, and Ferda Ofli. 2021.
\newblock \href
  {https://ojs.aaai.org/index.php/ICWSM/article/download/18116/17919/21611}
  {Humaid: Human-annotated disaster incidents data from twitter with deep
  learning benchmarks}.
\newblock In \emph{Proceedings of the Fifteenth International AAAI Conference
  on Web and Social Media, ICWSM. AAAI Press}, pages 933--942.

\bibitem[{Bamler and Mandt(2017)}]{bamler2017dynamic}
Robert Bamler and Stephan Mandt. 2017.
\newblock \href {http://proceedings.mlr.press/v70/bamler17a.html} {Dynamic word
  embeddings}.
\newblock In \emph{Proceedings of the 34th International Conference on Machine
  Learning, {ICML} 2017, Sydney, NSW, Australia, 6-11 August 2017}, volume~70
  of \emph{Proceedings of Machine Learning Research}, pages 380--389. {PMLR}.

\bibitem[{Bjerva et~al.(2020)Bjerva, Kouw, and Augenstein}]{bjerva2019}
Johannes Bjerva, Wouter Kouw, and Isabelle Augenstein. 2020.
\newblock \href {https://ojs.aaai.org//index.php/AAAI/article/view/6240} {Back
  to the future --- sequential alignment of text representations}.
\newblock In \emph{Proceedings of the AAAI Conference on Artificial
  Intelligence}, volume~34, pages 7440--7447, United States. AAAI Press.

\bibitem[{Del~Tredici et~al.(2019)Del~Tredici, Fern{\'a}ndez, and
  Boleda}]{del-tredici-etal-2019-short}
Marco Del~Tredici, Raquel Fern{\'a}ndez, and Gemma Boleda. 2019.
\newblock \href {https://doi.org/10.18653/v1/N19-1210} {Short-term meaning
  shift: A distributional exploration}.
\newblock In \emph{Proceedings of the 2019 Conference of the North {A}merican
  Chapter of the Association for Computational Linguistics: Human Language
  Technologies, Volume 1 (Long and Short Papers)}, pages 2069--2075,
  Minneapolis, Minnesota. Association for Computational Linguistics.

\bibitem[{Devlin et~al.(2019)Devlin, Chang, Lee, and
  Toutanova}]{devlin-etal-2019-bert}
Jacob Devlin, Ming-Wei Chang, Kenton Lee, and Kristina Toutanova. 2019.
\newblock \href {https://doi.org/10.18653/v1/N19-1423} {{BERT}: Pre-training of
  deep bidirectional transformers for language understanding}.
\newblock In \emph{Proceedings of the 2019 Conference of the North {A}merican
  Chapter of the Association for Computational Linguistics: Human Language
  Technologies, Volume 1 (Long and Short Papers)}, pages 4171--4186,
  Minneapolis, Minnesota. Association for Computational Linguistics.

\bibitem[{Dhingra et~al.(2022)Dhingra, Cole, Eisenschlos, Gillick, Eisenstein,
  and Cohen}]{dhingra2021time}
Bhuwan Dhingra, Jeremy~R. Cole, Julian~Martin Eisenschlos, Daniel Gillick,
  Jacob Eisenstein, and William~W. Cohen. 2022.
\newblock \href {https://doi.org/10.1162/tacl_a_00459} {{Time-Aware Language
  Models as Temporal Knowledge Bases}}.
\newblock \emph{Transactions of the Association for Computational Linguistics},
  10:257--273.

\bibitem[{Eisenstein(2013)}]{eisenstein2013}
Jacob Eisenstein. 2013.
\newblock \href {https://www.aclweb.org/anthology/N13-1037} {What to do about
  bad language on the internet}.
\newblock In \emph{Proceedings of the 2013 Conference of the North {A}merican
  Chapter of the Association for Computational Linguistics: Human Language
  Technologies}, pages 359--369, Atlanta, Georgia. Association for
  Computational Linguistics.

\bibitem[{Florio et~al.(2020)Florio, Basile, Polignano, Basile, and
  Patti}]{app10124180}
Komal Florio, Valerio Basile, Marco Polignano, Pierpaolo Basile, and Viviana
  Patti. 2020.
\newblock \href {https://doi.org/10.3390/app10124180} {{Time of Your Hate: The
  Challenge of Time in Hate Speech Detection on Social Media}}.
\newblock \emph{Applied Sciences}, 10(12).

\bibitem[{Ganin et~al.(2016)Ganin, Ustinova, Ajakan, Germain, Larochelle,
  Laviolette, Marchand, and Lempitsky}]{ganin2016domain}
Yaroslav Ganin, Evgeniya Ustinova, Hana Ajakan, Pascal Germain, Hugo
  Larochelle, Fran{\c{c}}ois Laviolette, Mario Marchand, and Victor Lempitsky.
  2016.
\newblock \href {https://www.jmlr.org/papers/volume17/15-239/15-239.pdf}
  {Domain-adversarial training of neural networks}.
\newblock \emph{The journal of machine learning research}, 17(1):2096--2030.

\bibitem[{Golder and Macy(2011)}]{golder2011}
Scott~A. Golder and Michael~W. Macy. 2011.
\newblock \href {https://doi.org/10.1126/science.1202775} {Diurnal and seasonal
  mood vary with work, sleep, and daylength across diverse cultures}.
\newblock \emph{Science}, 333(6051):1878--1881.

\bibitem[{Gonen et~al.(2020)Gonen, Jawahar, Seddah, and
  Goldberg}]{gonen-etal-2020-simple}
Hila Gonen, Ganesh Jawahar, Djam{\'e} Seddah, and Yoav Goldberg. 2020.
\newblock \href {https://doi.org/10.18653/v1/2020.acl-main.51} {Simple,
  interpretable and stable method for detecting words with usage change across
  corpora}.
\newblock In \emph{Proceedings of the 58th Annual Meeting of the Association
  for Computational Linguistics}, pages 538--555, Online. Association for
  Computational Linguistics.

\bibitem[{Gururangan et~al.(2020)Gururangan, Marasovi{\'c}, Swayamdipta, Lo,
  Beltagy, Downey, and Smith}]{gururangan-etal-2020-dont}
Suchin Gururangan, Ana Marasovi{\'c}, Swabha Swayamdipta, Kyle Lo, Iz~Beltagy,
  Doug Downey, and Noah~A. Smith. 2020.
\newblock \href {https://doi.org/10.18653/v1/2020.acl-main.740} {Don{'}t stop
  pretraining: Adapt language models to domains and tasks}.
\newblock In \emph{Proceedings of the 58th Annual Meeting of the Association
  for Computational Linguistics}, pages 8342--8360, Online. Association for
  Computational Linguistics.

\bibitem[{Hamilton et~al.(2016{\natexlab{a}})Hamilton, Leskovec, and
  Jurafsky}]{hamilton2016}
William~L. Hamilton, Jure Leskovec, and Dan Jurafsky. 2016{\natexlab{a}}.
\newblock \href {https://doi.org/10.18653/v1/P16-1141} {Diachronic word
  embeddings reveal statistical laws of semantic change}.
\newblock In \emph{Proceedings of the 54th Annual Meeting of the Association
  for Computational Linguistics (Volume 1: Long Papers)}, pages 1489--1501,
  Berlin, Germany. Association for Computational Linguistics.

\bibitem[{Hamilton et~al.(2016{\natexlab{b}})Hamilton, Leskovec, and
  Jurafsky}]{hamilton-etal-2016-diachronic}
William~L. Hamilton, Jure Leskovec, and Dan Jurafsky. 2016{\natexlab{b}}.
\newblock \href {https://doi.org/10.18653/v1/P16-1141} {Diachronic word
  embeddings reveal statistical laws of semantic change}.
\newblock In \emph{Proceedings of the 54th Annual Meeting of the Association
  for Computational Linguistics (Volume 1: Long Papers)}, pages 1489--1501,
  Berlin, Germany. Association for Computational Linguistics.

\bibitem[{Hofmann et~al.(2021)Hofmann, Pierrehumbert, and
  Sch{\"u}tze}]{hofmann-etal-2021-dynamic}
Valentin Hofmann, Janet Pierrehumbert, and Hinrich Sch{\"u}tze. 2021.
\newblock \href {https://doi.org/10.18653/v1/2021.acl-long.542} {Dynamic
  contextualized word embeddings}.
\newblock In \emph{Proceedings of the 59th Annual Meeting of the Association
  for Computational Linguistics and the 11th International Joint Conference on
  Natural Language Processing (Volume 1: Long Papers)}, pages 6970--6984,
  Online. Association for Computational Linguistics.

\bibitem[{Jaidka et~al.(2018)Jaidka, Chhaya, and
  Ungar}]{jaidka-etal-2018-diachronic}
Kokil Jaidka, Niyati Chhaya, and Lyle Ungar. 2018.
\newblock \href {https://doi.org/10.18653/v1/P18-2032} {Diachronic degradation
  of language models: Insights from social media}.
\newblock In \emph{Proceedings of the 56th Annual Meeting of the Association
  for Computational Linguistics (Volume 2: Short Papers)}, pages 195--200,
  Melbourne, Australia. Association for Computational Linguistics.

\bibitem[{Jawahar and Seddah(2019)}]{jawahar-seddah-2019-contextualized}
Ganesh Jawahar and Djam{\'e} Seddah. 2019.
\newblock \href {https://doi.org/10.18653/v1/W19-4705} {Contextualized
  diachronic word representations}.
\newblock In \emph{Proceedings of the 1st International Workshop on
  Computational Approaches to Historical Language Change}, pages 35--47,
  Florence, Italy. Association for Computational Linguistics.

\bibitem[{Kulkarni et~al.(2015)Kulkarni, Al-Rfou, Perozzi, and
  Skiena}]{kulkarni2015}
Vivek Kulkarni, Rami Al-Rfou, Bryan Perozzi, and Steven Skiena. 2015.
\newblock \href {http://viveksck.github.io/langchangetrack/data/kulkarni.pdf}
  {Statistically significant detection of linguistic change}.
\newblock In \emph{Proceedings of the 24th International Conference on World
  Wide Web}, pages 1384--1397, Florence, Italy. Association for Computing
  Machinery.

\bibitem[{Kulkarni et~al.(2021)Kulkarni, Mishra, and
  Haghighi}]{kulkarni-etal-2021-lmsoc-approach}
Vivek Kulkarni, Shubhanshu Mishra, and Aria Haghighi. 2021.
\newblock \href {https://doi.org/10.18653/v1/2021.findings-emnlp.254} {{LMSOC}:
  An approach for socially sensitive pretraining}.
\newblock In \emph{Findings of the Association for Computational Linguistics:
  EMNLP 2021}, pages 2967--2975, Punta Cana, Dominican Republic. Association
  for Computational Linguistics.

\bibitem[{Kutuzov et~al.(2018)Kutuzov, {\O}vrelid, Szymanski, and
  Velldal}]{kutuzov-etal-2018-diachronic}
Andrey Kutuzov, Lilja {\O}vrelid, Terrence Szymanski, and Erik Velldal. 2018.
\newblock \href {https://aclanthology.org/C18-1117} {Diachronic word embeddings
  and semantic shifts: a survey}.
\newblock In \emph{Proceedings of the 27th International Conference on
  Computational Linguistics}, pages 1384--1397, Santa Fe, New Mexico, USA.
  Association for Computational Linguistics.

\bibitem[{Lazaridou et~al.(2021)Lazaridou, Kuncoro, Gribovskaya, Agrawal,
  Liska, Terzi, Gimenez, de~Masson~d'Autume, Ko{\v{c}}isk{\'{y}}, Ruder,
  Yogatama, Cao, Young, and Blunsom}]{lazaridou2021mind}
Angeliki Lazaridou, Adhiguna Kuncoro, Elena Gribovskaya, Devang Agrawal, Adam
  Liska, Tayfun Terzi, Mai Gimenez, Cyprien de~Masson~d'Autume,
  Tom{\'{a}}{\v{s}} Ko{\v{c}}isk{\'{y}}, Sebastian Ruder, Dani Yogatama, Kris
  Cao, Susannah Young, and Phil Blunsom. 2021.
\newblock \href {https://openreview.net/forum?id=73OmmrCfSyy} {{Mind the Gap:
  Assessing Temporal Generalization in Neural Language Models}}.
\newblock In \emph{Advances in Neural Information Processing Systems}.

\bibitem[{Loshchilov and Hutter(2019)}]{loshchilov2018decoupled}
Ilya Loshchilov and Frank Hutter. 2019.
\newblock \href {https://openreview.net/forum?id=Bkg6RiCqY7} {Decoupled weight
  decay regularization}.
\newblock In \emph{7th International Conference on Learning Representations,
  {ICLR} 2019, New Orleans, LA, USA, May 6-9, 2019}. OpenReview.net.

\bibitem[{Loureiro et~al.(2022{\natexlab{a}})Loureiro, Barbieri, Neves,
  Espinosa~Anke, and Camacho-collados}]{loureiro2022timelms}
Daniel Loureiro, Francesco Barbieri, Leonardo Neves, Luis Espinosa~Anke, and
  Jose Camacho-collados. 2022{\natexlab{a}}.
\newblock \href {https://doi.org/10.18653/v1/2022.acl-demo.25} {{T}ime{LM}s:
  Diachronic language models from {T}witter}.
\newblock In \emph{Proceedings of the 60th Annual Meeting of the Association
  for Computational Linguistics: System Demonstrations}, pages 251--260,
  Dublin, Ireland. Association for Computational Linguistics.

\bibitem[{Loureiro et~al.(2022{\natexlab{b}})Loureiro, D'Souza, Muhajab, White,
  Wong, Anke, Neves, Barbieri, and Camacho-Collados}]{tempowic_coling22}
Daniel Loureiro, Aminette D'Souza, Areej~Nasser Muhajab, Isabella~A. White,
  Gabriel Wong, Luis~Espinosa Anke, Leonardo Neves, Francesco Barbieri, and
  Jose Camacho-Collados. 2022{\natexlab{b}}.
\newblock \href {https://doi.org/10.48550/ARXIV.2209.07216} {Tempowic: An
  evaluation benchmark for detecting meaning shift in social media}.

\bibitem[{Luu et~al.(2022)Luu, Khashabi, Gururangan, Mandyam, and
  Smith}]{luu2021time}
Kelvin Luu, Daniel Khashabi, Suchin Gururangan, Karishma Mandyam, and Noah~A.
  Smith. 2022.
\newblock \href {https://doi.org/10.18653/v1/2022.naacl-main.435} {Time waits
  for no one! analysis and challenges of temporal misalignment}.
\newblock In \emph{Proceedings of the 2022 Conference of the North American
  Chapter of the Association for Computational Linguistics: Human Language
  Technologies}, pages 5944--5958, Seattle, United States. Association for
  Computational Linguistics.

\bibitem[{Martinc et~al.(2020)Martinc, Kralj~Novak, and
  Pollak}]{martinc-etal-2020-leveraging}
Matej Martinc, Petra Kralj~Novak, and Senja Pollak. 2020.
\newblock \href {https://aclanthology.org/2020.lrec-1.592} {Leveraging
  contextual embeddings for detecting diachronic semantic shift}.
\newblock In \emph{Proceedings of the Twelfth Language Resources and Evaluation
  Conference}, pages 4811--4819, Marseille, France. European Language Resources
  Association.

\bibitem[{Ni et~al.(2019)Ni, Li, and McAuley}]{amazon-review}
Jianmo Ni, Jiacheng Li, and Julian McAuley. 2019.
\newblock \href {https://doi.org/10.18653/v1/D19-1018} {Justifying
  recommendations using distantly-labeled reviews and fine-grained aspects}.
\newblock In \emph{Proceedings of the 2019 Conference on Empirical Methods in
  Natural Language Processing and the 9th International Joint Conference on
  Natural Language Processing (EMNLP-IJCNLP)}, pages 188--197, Hong Kong,
  China. Association for Computational Linguistics.

\bibitem[{Olteanu et~al.(2015)Olteanu, Vieweg, and Castillo}]{olteanu2015}
Alexandra Olteanu, Sarah Vieweg, and Carlos Castillo. 2015.
\newblock \href {https://doi.org/10.1145/2675133.2675242} {What to expect when
  the unexpected happens: Social media communications across crises}.
\newblock In \emph{Proceedings of the 18th ACM Conference on Computer Supported
  Cooperative Work and Social Computing}, CSCW '15, page 994–1009, New York,
  New York. Association for Computing Machinery.

\bibitem[{Paszke et~al.(2019)Paszke, Gross, Massa, Lerer, Bradbury, Chanan,
  Killeen, Lin, Gimelshein, Antiga, Desmaison, K{\"{o}}pf, Yang, DeVito,
  Raison, Tejani, Chilamkurthy, Steiner, Fang, Bai, and Chintala}]{pytorch2019}
Adam Paszke, Sam Gross, Francisco Massa, Adam Lerer, James Bradbury, Gregory
  Chanan, Trevor Killeen, Zeming Lin, Natalia Gimelshein, Luca Antiga, Alban
  Desmaison, Andreas K{\"{o}}pf, Edward~Z. Yang, Zachary DeVito, Martin Raison,
  Alykhan Tejani, Sasank Chilamkurthy, Benoit Steiner, Lu~Fang, Junjie Bai, and
  Soumith Chintala. 2019.
\newblock \href
  {https://proceedings.neurips.cc/paper/2019/hash/bdbca288fee7f92f2bfa9f7012727740-Abstract.html}
  {Pytorch: An imperative style, high-performance deep learning library}.
\newblock In \emph{Advances in Neural Information Processing Systems 32: Annual
  Conference on Neural Information Processing Systems 2019, NeurIPS 2019,
  December 8-14, 2019, Vancouver, BC, Canada}, pages 8024--8035.

\bibitem[{Ramponi and Plank(2020)}]{ramponi-plank-2020-neural}
Alan Ramponi and Barbara Plank. 2020.
\newblock \href {https://doi.org/10.18653/v1/2020.coling-main.603} {Neural
  unsupervised domain adaptation in {NLP}{---}{A} survey}.
\newblock In \emph{Proceedings of the 28th International Conference on
  Computational Linguistics}, pages 6838--6855, Barcelona, Spain (Online).
  International Committee on Computational Linguistics.

\bibitem[{Reimers and Gurevych(2019)}]{reimers-2019-sentence-bert}
Nils Reimers and Iryna Gurevych. 2019.
\newblock \href {https://arxiv.org/abs/1908.10084} {Sentence-bert: Sentence
  embeddings using siamese bert-networks}.
\newblock In \emph{Proceedings of the 2019 Conference on Empirical Methods in
  Natural Language Processing}. Association for Computational Linguistics.

\bibitem[{Reynolds and Seeger(2005)}]{reynolds2005}
Barbara Reynolds and Mathew~W. Seeger. 2005.
\newblock \href {https://doi.org/10.1080/10810730590904571} {Crisis and
  emergency risk communication as an integrative model}.
\newblock \emph{Journal of Health Communication}, 10(1):43--55.

\bibitem[{R{\"o}ttger and
  Pierrehumbert(2021)}]{rottger-pierrehumbert-2021-temporal-adaptation}
Paul R{\"o}ttger and Janet Pierrehumbert. 2021.
\newblock \href {https://doi.org/10.18653/v1/2021.findings-emnlp.206} {Temporal
  adaptation of {BERT} and performance on downstream document classification:
  Insights from social media}.
\newblock In \emph{Findings of the Association for Computational Linguistics:
  EMNLP 2021}, pages 2400--2412, Punta Cana, Dominican Republic. Association
  for Computational Linguistics.

\bibitem[{Rudolph and Blei(2018)}]{10.1145/3178876.3185999}
Maja Rudolph and David Blei. 2018.
\newblock \href {https://doi.org/10.1145/3178876.3185999} {Dynamic embeddings
  for language evolution}.
\newblock In \emph{Proceedings of the 2018 World Wide Web Conference}, WWW '18,
  page 1003–1011, Republic and Canton of Geneva, CHE. International World
  Wide Web Conferences Steering Committee.

\bibitem[{Sandhaus(2008)}]{nyt-2008}
Evan Sandhaus. 2008.
\newblock \href {https://doi.org/10.35111/77ba-9x74} {The new york times
  annotated corpus ldc2008t19.}
\newblock In \emph{Linguistic Data Consortium, 2008}. Linguistic Data
  Consortium.

\bibitem[{Sommerauer and Fokkens(2019)}]{sommerauer2019}
Pia Sommerauer and Antske Fokkens. 2019.
\newblock \href {https://doi.org/10.18653/v1/W19-4728} {Conceptual change and
  distributional semantic models: an exploratory study on pitfalls and
  possibilities}.
\newblock In \emph{Proceedings of the 1st International Workshop on
  Computational Approaches to Historical Language Change}, pages 223--233,
  Florence, Italy. Association for Computational Linguistics.

\bibitem[{Stowe et~al.(2018)Stowe, Anderson, Palmer, Palen, and
  Anderson}]{stowe2018}
Kevin Stowe, Jennings Anderson, Martha Palmer, Leysia Palen, and Ken Anderson.
  2018.
\newblock \href {https://doi.org/10.18653/v1/W18-3512} {Improving
  classification of {T}witter behavior during hurricane events}.
\newblock In \emph{Proceedings of the Sixth International Workshop on Natural
  Language Processing for Social Media}, pages 67--75, Melbourne, Australia.
  Association for Computational Linguistics.

\bibitem[{Wang et~al.(2014)Wang, Zhang, Feng, and
  Chen}]{Wang_Zhang_Feng_Chen_2014}
Zhen Wang, Jianwen Zhang, Jianlin Feng, and Zheng Chen. 2014.
\newblock \href {https://ojs.aaai.org/index.php/AAAI/article/view/8870}
  {Knowledge graph embedding by translating on hyperplanes}.
\newblock \emph{Proceedings of the AAAI Conference on Artificial Intelligence},
  28(1).

\bibitem[{Wolf et~al.(2020)Wolf, Debut, Sanh, Chaumond, Delangue, Moi, Cistac,
  Rault, Louf, Funtowicz, Davison, Shleifer, von Platen, Ma, Jernite, Plu, Xu,
  Le~Scao, Gugger, Drame, Lhoest, and Rush}]{wolf-etal-2020-transformers}
Thomas Wolf, Lysandre Debut, Victor Sanh, Julien Chaumond, Clement Delangue,
  Anthony Moi, Pierric Cistac, Tim Rault, Remi Louf, Morgan Funtowicz, Joe
  Davison, Sam Shleifer, Patrick von Platen, Clara Ma, Yacine Jernite, Julien
  Plu, Canwen Xu, Teven Le~Scao, Sylvain Gugger, Mariama Drame, Quentin Lhoest,
  and Alexander Rush. 2020.
\newblock \href {https://doi.org/10.18653/v1/2020.emnlp-demos.6} {Transformers:
  State-of-the-art natural language processing}.
\newblock In \emph{Proceedings of the 2020 Conference on Empirical Methods in
  Natural Language Processing: System Demonstrations}, pages 38--45, Online.
  Association for Computational Linguistics.

\bibitem[{Yang et~al.(2013)Yang, Chung, Lin, Lee, Chen, Wood, Kavanaugh,
  Sheetz, Shoemaker, and Fox}]{yang2013}
Seungwon Yang, Haeyong Chung, Xiao Lin, Sunshin Lee, Liangzhe Chen, Andrew
  Wood, Andrea~L. Kavanaugh, Steven~D. Sheetz, Donald~J. Shoemaker, and
  Edward~A. Fox. 2013.
\newblock \href {http://idl.iscram.org/files/yang/2013/1122_Yang_etal2013.pdf}
  {Phasevis: What, when, where, and who in visualizing the four phases of
  emergency management through the lens of social media}.
\newblock In \emph{Proceedings of the 10th International ISCRAM Conference},
  pages 912--917, Baden-Baden, Germany.

\end{thebibliography}
\bibliographystyle{acl_natbib}

\appendix

\section{Appendix: Data}
\label{sec:appendixa}
\subsection{Data Statistics}

The \sandy{} dataset spans 18 days with 23k tweets. 
The \humaid{} datasets range from 560 to 9399 tweets, from 1 to 81 days. 
The \clex{} datasets range from 1000 to 1442 tweets, over 7 to 56 days.
In Table~\ref{tab:clex_counts} and \ref{tab:humaid_counts} we show the dataset statistics for the \clex{} datasets and \humaid{} datasets, respectively.
Note that the Typhoon Pablo event from the original \clex{} dataset had only seven unlabelled tweets that could be successfully recovered: we therefore remove it from all experiments.

\begin{table*}[ht]
	\centering
	\small
	\begin{tabular}{l|l|r|r}
		\multicolumn{4}{c}{\textbf{Progressive Events}} \\
		\hline
		Event & Dates (MM.DD.YY) & Total Days & Tweets \\
		\hline
		Colorado Floods (2013) & 09.08.13 - 10.01.13 & 19 &  1,231 \\
		Sardinia Floods (2013) & 11.16.13 - 11.28.13 & 13 & 824 \\ 
		Philipinnes Floods (2012) & 08.07.12 - 08.15.12 & 13 & 1,341 \\
		Alberta Floods (2013) & 06.20.13 - 07.16.13 & 24 & 4,040 \\
		Manila Flood (2013) & 08.17.13 - 08.27.13 & 11 &  1,068 \\
		Queensland loods (2013) & 01.17.13 - 02.05.13 & 19 & 727 \\ 
		Typhoon Yolanda (2013) & 05.11.13 - 12.30.13 & 53 &  253 \\      
		Australia bushfire (2013) & 10.12.13 - 11.03.13 & 22 &  1,244 \\            
		Colorado wildfires (2012) & 06.08.12 - 07.08.12 & 31 &  2,901 \\
		Singapore haze (2013) & 06.14.13 - 07.04.13 & 18 & 1,572 \\
		\hline
		\multicolumn{4}{c}{\textbf{Instantaneous Events}} \\
		\hline
		Italy earthquakes (2012) & 05.18.12 - 06.14.12 & 28 & 5,219 \\
		Costa Rica earthquake (2012) & 09.05.12 - 09.21.12 & 18 & 1,641 \\
		Bohol earthquake (2013) & 10.14.13 - 10.25.13 & 12 & 1,131 \\
		Guatemala earthquake (2012) & 11.06.12 - 11.25.12 & 20 & 2,233 \\
		LA airport shootings (2013) & 11.01.13 - 11.12.13 & 12 & 1,737 \\
		Boston bombings (2013) & 04.15.13 - 06.11.13 & 46 & 81,172 \\
		West Texas explosion (2013) & 04.18.13 - 05.15.13 & 27 & 8,152 \\
		Venezuela refinery explosion (2012) & 12.08.24 - 12.09.05 & 13 & 2,007 \\
		Brazil nightclub fire (2013) & 01.27.13 - 02.11.13 & 16 & 2,644 \\
		Savar building collapse (2013) & 04.23.13 - 06.01.13 & 39 & 2,646 \\
		Spain train crash (2013) & 07.24.13 - 08.07.13 & 14 & 2,288 \\
		Lac Megantic train crash (2013) & 07.06.12 - 07.26.12 & 21 & 1,755 \\
		NY train crash (2013) & 12.01.13 - 12.08.13 & 9 & 667 \\
		Glasgow helicopter crash (2013) & 11.29.13 - 12.29.13 & 30 & 1,541 \\
		Russia meteor (2013) & 02.14.13 - 03.05.13 & 19 & 4,289 \\
	\end{tabular}
	\caption{
		Summary of the \clex datasets. The \textit{progressive} and \textit{instantaneous} splits were done manually based on the type of crisis event.
	}
	\label{tab:clex_counts}
\end{table*}

\begin{table*}[ht]
	\centering
	\small
	\begin{tabular}{l|l|r|r}
		\multicolumn{4}{c}{\textbf{Progressive Events}} \\
		\hline
		Event (Year) & Dates (MM.DD.YY) & Total Days & Nr. Tweets \\
		\hline
		Canada Wildfires (2016) & 17.04.16 - 25.12.16 & 253 & 2,258 \\
		Hurricane Matthew (2016) & 04.10.16 - 05.12.16 & 74 & 1,659 \\
		Sri Lanka Floods (2017) & 31.05.17 - 03.07.17 & 34 & 575 \\
		Hurricane Harvey (2017) & 17.08.17 - 19.09.17 & 34 & 9,164 \\
		Hurricane Irma (2017) & 06.09.17 - 21.09.17 & 16 & 9,467 \\
		Hurricane Maria (2017) & 16.09.17 - 02.10.17 & 17 & 7,328 \\
		Maryland Floods (2018) & 28.05.18 - 07.06.18 & 11 & 747 \\
		Greece Wildfires (2018) & 24.07.18 - 18.08.18 & 26 & 1,526 \\
		Kerala Floods (2018) & 17.08.18 - 12.09.18 & 27 & 8,056 \\
		Hurricane Florence (2018) & 11.09.18 - 17.11.18 & 68 & 6,359 \\
		California Wildfires (2018) & 10.11.18 - 07.12.18 & 28 & 7,444 \\
		Cyclone Idai (2019) & 15.03.19 - 16.04.19 & 33 & 3,944 \\
		Midwestern U.S. Floods (2019) & 25.03.19 - 03.04.19 & 26 & 1,930 \\
		Hurricane Dorian (2019) & 30.08.19 - 02.09.19 & 4 & 7,660 \\
		\hline
		\multicolumn{4}{c}{\textbf{Instantaneous Events}} \\
		\hline
		Ecuador Earthquake (2016) & 17.04.16 - 25.12.16 & 253 & 1,594 \\
		Italy Earthquake (2016) & 24.08.16 - 29.08.16 & 6 & 1,240 \\
		Kaikoura Earthquake (2016) & 01.09.16 - 22.11.16 & 83 & 2,217 \\
		Mexico Earthquake (2017) & 20.09.17 - 06.10.17 & 17 & 2,036 \\
		Pakistan Earthquake (2019) & 24.09.19 - 26.09.19 & 3 & 1,991 \\
	\end{tabular}
	\caption{
		Summary of the \humaid{} datasets. The \textit{progressive} and \textit{instantaneous} splits were done manually based on the type of crisis event.
	}
	\label{tab:humaid_counts}
\end{table*}

\subsection{Detailed Results for \clex{} and \humaid{}}
\label{app:results}
In Tables \ref{tab:detail_eval_clex} and \ref{tab:detail_eval_humaid} we provide the detailed evaluation results of the proposed approaches on \clex{} and \humaid{}.


\begin{table*}[!ht]
    \small
    \begin{center}
    \scalebox{0.85}{
        \begin{tabular}{l c c c c c c c c c c c c}
            \toprule
            \multirow{2}{*}{Event} & & \multicolumn{11}{c}{\humaid} \\
            
            \cmidrule(lr){2-13} 
            
            & \multicolumn{2}{c}{BERT} & \multicolumn{2}{c}{BERT+TM} & \multicolumn{2}{c}{BERT+SEP} & \multicolumn{2}{c}{BERT+DCWE} & \multicolumn{2}{c}{BERT+TAPH} & \multicolumn{2}{c}{BERT+TDA} \\
            
            \cmidrule(lr){2-3} \cmidrule(lr){4-5} \cmidrule(lr){6-7} \cmidrule(lr){8-9} \cmidrule(lr){10-11} \cmidrule(lr){12-13} \\
			
			& \textsc{cont} & \textsc{temp} & \textsc{cont} & \textsc{temp} & \textsc{cont} & \textsc{temp} & \textsc{cont} & \textsc{temp} & \textsc{cont} & \textsc{temp} & \textsc{cont} & \textsc{temp} \\
			
			\hline
			\multicolumn{13}{c}{\bf Progressive Events} \\
			\hline 
			
Colorado Floods (2013) &	0.309 &	0.309 & 0.309 & 0.309 & 0.309 & 0.309  &	0.309 &	0.309  & 0.309 &	0.309 &  0.309 &	0.309 \\
			
Sardinia Floods (2013) &	0.255	&  0.315 & 0.310 & 0.287 & 0.239 & 0.285  &	0.179 &	0.298  & 0.179 & 0.211 &  0.299 &	0.288 \\
		    
Philipinnes Floods (2012 &	0.276 &  0.270 & 0.307 & 0.269 & 0.213 & 0.278  & 0.213 &	0.213 & 0.213 & 0.213 &	0.213 &	0.269 \\
		    
Alberta Floods (2013) &  0.314 &	0.202 & 0.307 & 0.200 & 0.300 & 0.202  & 0.202 & 0.202 & 0.202 & 0.202 & 0.296 &	0.202 \\
		    
Manila Floods (2013) & 0.369 & 	0.369 & 0.367 & 0.366 & 0.337 & 0.372  &	0.190 &	0.350 & 0.308 & 0.355 &  0.380 &	 0.374 \\
		    
Queensland Floods (2013) &	0.423 &	0.353 & 0.486 & 0.342 & 0.361 & 0.331  & 0.374 & 0.351 & 0.318 & 0.314 & 	 0.472 &  0.355 \\
		    
Typhoon Yolanda (2013) & 0.211 & 0.211 & 0.235 & 0.260 & 0.317 & 0.399  & 0.211 & 0.211 & 0.211 & 0.211 & 0.211 & 0.211 \\
		    
Australia Bushfire (2013) &	0.447 &	0.450 & 0.583 & 0.585 & 0.449 & 0.522  &	0.426 &	0.421 & 0.422 & 0.461  &  0.577 &  0.547 \\
		    
Colorado Wildfires (2012) &  0.569 &  0.370 & 0.584 & 0.370 & 0.541 & 0.335  & 0.533 & 0.370 & 0.446 & 0.330 &	0.567 &	0.222 \\
		    
Singapore Haze (2013) &	 0.363 &	0.348 & 0.360 & 0.340 & 0.352 & 0.344  & 0.357 & 0.332  & 0.361 & 0.349 &	0.360	 &  0.351 \\

			\hline
			\multicolumn{13}{c}{\bf Instantaneous Events} \\
			\hline
			
Italy Earthquakes (2012) &	 0.332 &	 0.321& 0.316 & 0.285 & 0.331 & 0.304  &	0.287 &	0.267 & 0.274 & 0.316 & 0.326 &	0.318 \\
			
Costa Rica Earthquake (2012) & 	 0.582 &	 0.240 & 0.564 & 0.132 & 0.603 & 0.102  & 0.554 & 0.102 & 0.537 & 0.102  &	0.543 &	0.102 \\
		    
Bohol Earthquake (2013) &	 0.585 &	 0.579 & 0.566 & 0.566 & 0.574 & 0.568 & 0.569 & 0.574 & 0.574 & 0.571 & 	0.582 &	0.577 \\
		    
Guatemala Earthquake (2012) &  0.568 &  0.484 & 0.401 & 0.437 & 0.274 & 0.274  &	0.425 &	0.274 & 0.274 & 0.274 &  0.474 & 0.434 \\
		    
LA Airport Shootings (2013) &   0.534 &  0.475 & 0.518 & 0.465 & 0.210 & 0.378  &	0.376 &	0.312  & 0.309 & 0.192  &	0.356 &	0.382 \\
			
Boston Bombings (2013) & 0.358 & 0.340 & 0.362 & 0.349 & 0.360 & 0.356  &  0.378 & 0.300 & 0.363 &	0.352  & 0.354 &	 0.361 \\

West Texas Explosion (2013) &  0.411 & 0.398 & 0.405 & 0.396 & 0.412 & 0.396  & 0.396 &	0.392 & 0.407 & 0.405  & 	0.407 &	 0.409  \\

Venezuela Refinery Explosion (2012) &  0.368 &  0.347 & 0.359 & 0.336 & 0.360 & 0.344  &	0.339 &	0.339  & 0.361 &	0.335  & 0.362 & 0.343 \\

Brazil Nightclub Fire (2013) & 0.426 &	 0.431 & 0.425 & 0.413 & 0.416 & 0.416  &	0.422 &	0.302  & 0.424 & 0.412  &	 0.431 &	0.315  \\

Savar Building Collapse (2013) & 0.426 &  0.352 & 0.424 & 0.347 & 0.404 & 0.348  & 0.413 & 0.227  &	0.411 &	0.180  &	0.413 &	0.200 \\

Spain Train Crash (2013) &  0.463 &	0.446 & 0.490 & 0.539 & 0.481 & 0.447  & 0.355 & 0.402  & 0.324 & 0.460  &	0.456 &  0.449  \\

Lac Megantic Train Crash (2013) & 0.319 & 0.318 & 0.326 & 0.318 & 0.310 & 0.174  &	0.289 &	0.270 & 0.301 & 0.210  &  0.325 &  0.319 \\

NY Train Crash (2013) &	0.490 &	0.573 & 0.520 & 0.566 & 0.490 & 0.565  &	0.490 &	0.490 & 0.490 & 0.490 & 0.490 &	0.742 \\

Glasgow Helicopter Crash (2013) & 0.554 & 0.292 & 0.527 & 0.290 & 0.543 & 0.292  &	0.502 &	0.309 &	0.390 &	0.298 &  0.491 &  0.321 \\

Russia Meteor (2013) & 0.392 &  0.412 & 0.372 & 0.339 & 0.412 & 0.412  &	0.296 &	0.324  & 0.324 & 0.305  &	0.321 &	0.316 \\

            \bottomrule
        \end{tabular}
        }
        \caption{
    		{\bf Results for the \clex{} datasets.} The \textit{progressive} and \textit{instantaneous} splits were done manually based on the type of crisis event.
    	}
    	\label{tab:detail_eval_clex}
    \end{center}
\end{table*}


\begin{table*}[!ht]
    \small
    \begin{center}
    \scalebox{0.85}{
        \begin{tabular}{l c c c c c c c c c c c c}
            \toprule
            \multirow{2}{*}{Event} & & \multicolumn{11}{c}{\humaid} \\
            
            \cmidrule(lr){2-13} 
            
            & \multicolumn{2}{c}{BERT} & \multicolumn{2}{c}{BERT+TM} & \multicolumn{2}{c}{BERT+SEP} & \multicolumn{2}{c}{BERT+DCWE} & \multicolumn{2}{c}{BERT+TAPH} & \multicolumn{2}{c}{BERT+TDA} \\
            
            \cmidrule(lr){2-3} \cmidrule(lr){4-5} \cmidrule(lr){6-7} \cmidrule(lr){8-9} \cmidrule(lr){10-11} \cmidrule(lr){12-13} \\
			
			& \textsc{cont} & \textsc{temp} & \textsc{cont} & \textsc{temp} & \textsc{cont} & \textsc{temp} & \textsc{cont} & \textsc{temp} & \textsc{cont} & \textsc{temp} & \textsc{cont} & \textsc{temp} \\
			
			\hline
			\multicolumn{13}{c}{\bf Progressive Events} \\
			\hline 
			
			Canada Wildfires (2016) & 0.419 &  0.414  & 0.420 & 0.410 & 0.353 & 0.319  & 0.235 & 0.244 &  0.248 & 0.249 &  0.376 & 0.367 \\
			
		    Hurricane Matthew (2016) & 0.355 & 0.261  & 0.396 & 0.257 & 0.317 & 0.131  & 0.317 & 0.131 & 0.335 & 0.118 & 0.369 & 0.273 \\
		    
		    Sri Lanka Floods (2017) & 0.092 & 0.092 & 0.092 & 0.092 & 0.092 & 0.092  & 0.092 & 0.092 & 0.092 & 0.092 & 0.092 & 0.092 \\
		    
		    Hurricane Harvey (2017) & 0.635 & 0.663 & 0.639 & 0.669 & 0.637 & 0.645  & 0.589 & 0.586 & 0.578 & 0.587 & 0.583 & 0.581 \\
		    
		    Hurricane Irma (2017) &  0.624 &  0.618 & 0.639 & 0.614 & 0.610 & 0.579  &	0.566 &	0.549 &	0.568 &	0.553 &	0.579 &	0.545 \\
		    
		    Hurricane Maria (2017) &  0.620 &  0.628  & 0.640 & 0.621 & 0.603 & 0.602  & 0.507 & 0.575 &  0.501 & 0.581 & 0.600 & 0.529 \\
		    
		    Maryland Floods (2018) & 0.183 & 0.147 & 0.197 & 0.141 & 0.173 & 0.077  & 0.208 &  0.166 &  0.188 & 0.101 & 0.198 & 0.155 \\
		    
		    Greece Wildfires (2018) &  0.216 &  0.199 & 0.219 & 0.198 & 0.212 & 0.106  & 0.214 & 0.104  & 0.214 & 0.106 &  0.232 & 0.176 \\
		    
		    Kerala Floods (2018) &  0.470 &  0.422 & 0.421 & 0.420 & 0.480 & 0.382 & 0.354 & 0.348 &  0.341 & 0.347  & 0.379 & 0.346 \\
		    
		    Hurricane Florence (2018) &	0.663 & 0.510 & 0.664 & 0.500 & 0.658 & 0.481  & 0.590 & 0.435  & 0.586 & 0.417 &  0.649 & 0.421 \\
		    
		    California Wildfires (2018) &  0.601	& 0.484 & 0.624 & 0.567 & 0.571 & 0.485  & 0.544	& 0.455	& 0.558 & 0.470 &  0.575 & 0.485 \\
		    
		    Cyclone Idai (2019) &  0.372 &  0.350 & 0.370 & 0.350 & 0.352 & 0.331  & 0.287 & 0.298 &  0.319 & 0.294 & 0.347 & 0.300 \\
		    
		    Midwestern U.S. Floods (2019) & 0.300 & 0.405 & 0.300 & 0.362 & 0.277 & 0.301  & 0.137 & 0.229 & 0.192 & 0.217 & 0.251 & 0.261 \\
		    
		    Hurricane Dorian (2019) &  0.560	&  0.554 & 0.550 & 0.559 & 0.568 & 0.557  & 0.553 & 0.527 & 0.552 & 0.470 & 0.554 & 0.533 \\
		    
			\hline
			\multicolumn{13}{c}{\bf Instantaneous Events} \\
			\hline
			
			Ecuador Earthquake (2016) & 0.298 &  0.186 & 0.310 & 0.158 & 0.260 & 0.148  & 0.309 & 0.163 &  0.236 & 0.146  &  0.311 & 0.182 \\
			
		    Italy Earthquake (2016) & 0.395 & 0.266  & 0.403 & 0.260 & 0.350 & 0.090  & 0.118 & 0.090 & 0.175 & 0.090 &  0.401 & 0.274  \\
		    
		    Kaikoura Earthquake (2016) & 0.434 &  0.353 & 0.426 & 0.350 & 0.283 & 0.251  & 0.205 & 0.164 & 0.229 & 0.196 & 0.484 & 0.266 \\
		    
		    Mexico Earthquake (2017) & 0.340 &  0.318 & 0.341 & 0.300 & 0.283 & 0.262  & 0.269 & 0.258 &  0.245 & 0.264 & 0.289 & 0.281 \\
		    
		    Pakistan Earthquake (2019) & 0.273 & 0.205 & 0.260 & 0.200 & 0.243 & 0.168  & 0.203 & 0.168 & 0.190 & 0.162  &  0.350 &  0.215 \\
			
            \bottomrule
        \end{tabular}       
        }
        \caption{
    		{\bf Results for the \humaid{} datasets.} The \textit{progressive} and \textit{instantaneous} splits were done manually based on the type of crisis event.
    	}
    	\label{tab:detail_eval_humaid}
    \end{center}
\end{table*}

\begin{table*}[ht]
\clearpage
\onecolumn
\begin{longtable}{ p{3in} p{3in} }
\toprule
\textbf{Tweet} & \textbf{Analysis} \\
\midrule
Rep. Michael Grimm says situation in Staten island is "another \textcolor{red}{Katrina} situation"  & TDA correctly identifies \emph{Katrina} as the name of the storm in the temporal context of hurricane {\it Sandy}, while other models fails.\\
\midrule
\textcolor{red}{\#queenscomingtogether} Eric Ulrich brought the keg donated by Russos on the bay. & Adversarial signal forces TDA to lean time-invariant embedding for the word \emph{\#queenscomingtogether}. 
\\
\midrule

\end{longtable}
\caption{\label{tab:discussion} Representative examples showing tweets that TDA model correctly classifies while other models fail. Refer Section \ref{sec:discussion} for details.}
\clearpage
\twocolumn
\end{table*}

\end{document}